\documentclass[letterpaper]{article}
\usepackage{aaai}
\usepackage{times}
\usepackage{helvet}
\usepackage{courier}
\usepackage{fancyvrb}
\usepackage{amsmath}
\usepackage{amsthm}
\usepackage{graphicx}
\usepackage[scriptsize]{subfigure}
\usepackage{float}
\newcommand{\commentout}[1]{}
\newcommand{\decsamples}{\ensuremath{\Delta}}
\begin{document}

\nocopyright

\title{Decision--Making with Complex Data Structures using Probabilistic Programming}
\author{Brian E. Ruttenberg \and Avi Pfeffer\\
Charles River Analytics\\
625 Mt Auburn St.\\
Cambridge MA 02140\\
\{bruttenberg, apfeffer\}@cra.com
}
\maketitle
\begin{abstract}
\begin{quote}
Existing decision--theoretic reasoning frameworks such as decision networks use simple data structures and processes. However, decisions are often made based on complex data structures, such as social networks and protein sequences, and rich processes involving those structures. We present a framework for representing decision problems with complex data structures using probabilistic programming, allowing probabilistic models to be created with programming language constructs such as data structures and control flow. We provide a way to use arbitrary data types with minimal effort from the user, and an approximate decision--making algorithm that is effective even when the information space is very large or infinite. Experimental results show our algorithm working on problems with very large information spaces.
\end{quote}
\end{abstract}

\section{Introduction}

Suppose you are a medical practitioner deciding the appropriate dose of a medication to administer to a patient. You have sequencing information about some protein in the patient; the protein is composed of DNA, which mutates at variable rates. Higher dosages are more effective for patients whose DNA mutates faster, but consequently also produce more intense side effects. 

\begin{figure}[t]
\centering
\includegraphics[width=.8\columnwidth]{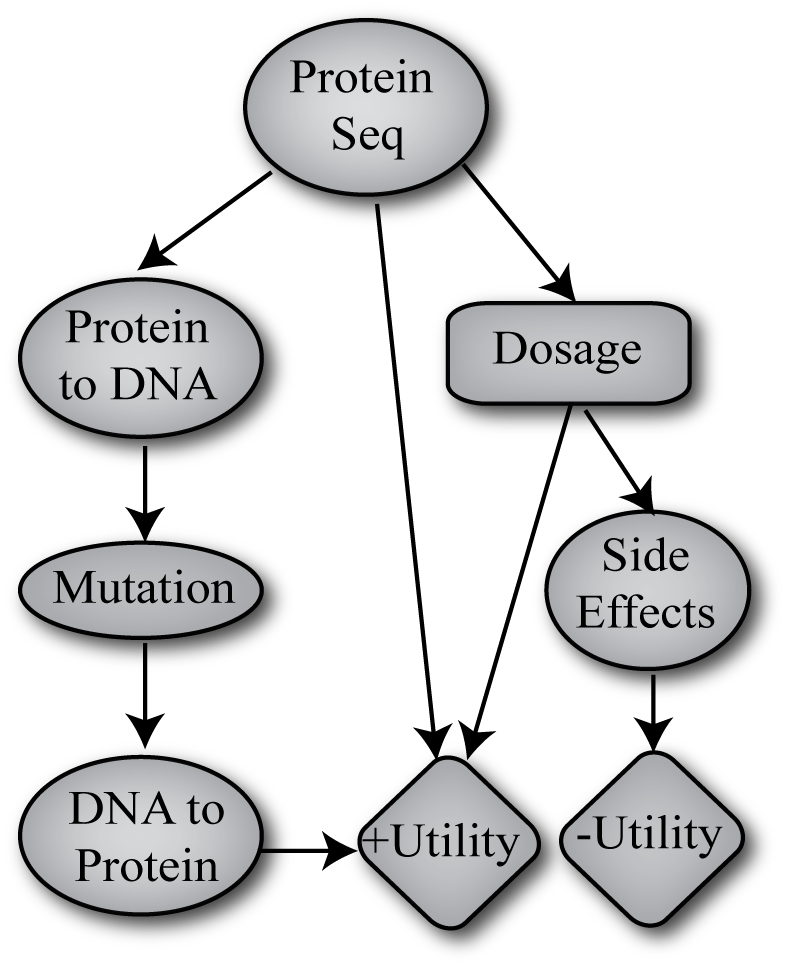}
\vspace{-4mm}
\caption{A decision network for a drug dosage scenario.\label{fig:DNA}}
\end{figure}

This problem fits into the classical decision--theoretic framework of decision--making based on partial state information and the uncertainty of the results upon making the actual decision. Indeed, it can be modeled naturally as a decision network or influence diagram~\cite{howard1984influence,howard2005influence}, as shown in Fig.~\ref{fig:DNA}. In this figure, ovals represent chance variables, the rectangle represents a decision, and the diamonds represent utility. The edge from the Protein Seq chance variable to the Dosage decision indicates that information about the protein sequence is available to the decision--maker at the time of making the dosage decision. Other edges indicate probabilistic dependency, as in Bayesian networks.

Current decision network frameworks cannot handle our problem. Current frameworks assume rudimentary data structures such as discrete enumerated sets or continuous variables, and the probabilistic relationships in the network are typically expressed as simple relationships such as conditional probability tables. In our problem, however, the protein and DNA sequence data structures are complex, as are the processes by which proteins map to DNA and the rate of DNA mutation. This presents two challenges: first, how do we represent decision problems with complex data structures, and second, how do we reason with them to create a policy that recommends the best decisions?

We address these challenges using probabilistic programming, which provides the ability to create probabilistic models using programming language constructs such as data structures and control flow. Probabilistic programming languages contain general purpose reasoning algorithms that can reason on all models written in the language. Probabilistic programming languages can naturally be extended with constructs denoting decisions, similar to the way decision networks extend Bayesian networks. By providing a general--purpose decision--making algorithm, all the benefits that probabilistic programming bestows upon standard probabilistic models are obtained for decision--theoretic models. In particular, we retain the ability to use complex data structure and processes with rich control flow, complete with decision--making algorithms that use them. Indeed, we provide a way to use arbitrary data types in probabilistic models with minimum effort from the user. We extend the publicly available Figaro language~\cite{figaro2012} to achieve this decision--making functionality.

Using complex data types as inputs to a decision presents a challenge because the space of possible inputs may be large or even infinite, which makes it difficult to precompute an optimal policy that specifies the best decision for every possible input. One possible approach is to compute the optimal policy online when a particular input is encountered. However, it may be too expensive to do this reasoning online for every possible input, and it may be preferable to compute a policy offline once and for all; indeed, computing a policy offline is a standard approach for decision problems. Furthermore, computing a policy online is not satisfactory when there are multiple decisions made in sequence, because earlier decisions depend on the policy of later decisions.

In the case of continuous inputs, a number of approximation approaches have been provided~\cite{shenoy2011inference,moral2001mixtures}, but these are not general--purpose solutions for arbitrary data types. We provide a sampling--based nearest neighbor algorithm that works for all data types, provided a distance function is defined for the type. Our algorithm is consistent such that in the infinite limit, it produces optimal decisions. Experimental results show that our algorithm produces small loss on several problems compared to alternative policies.

\section{Decision-Theoretic Probabilistic Programming}

\subsection{Figaro Programming Language}

Figaro is an open source, object--oriented, functional programming language implemented as a library in Scala. In Figaro, models are objects that define a stochastic or non--stochastic process. This means that models can be manipulated and operated on like traditional data structures, providing a natural means to create complex models that use inheritance, encapsulation and polymorphism. Figaro supports many models, such as directed or undirected graphical models, relational and recursive models, and open universe models. 

The basic unit in Figaro is the \texttt{Element[V]} class, parameterized by V, the output type of the element. At the most basic level, an element is an atomic unit that outputs some value based on a random process. For example, \texttt{Geometric(0.9)} is an \texttt{Element[Int]} that produces an integer output according to a geometric process whose parameter is 0.9.

In Figaro, elements can be passed to other elements as arguments, and a program typically consists of some number of element definitions. For example, a simple normal mixture model could be implemented as
\begin{Verbatim}[fontsize=\footnotesize]
 val n0 = Normal(0.0, 1.0)
 val n1 = Normal(5.0, 2.0)
 val mixProb = Flip(0.7)
 val mixModel = If(mixProb, n0, n1)
\end{Verbatim} 

Two key Figaro constructs that are essential to defining control flow in models are \texttt{Apply} and \texttt{Chain}.
\texttt{Apply} is a Figaro class that takes an \texttt{Element[T]} and an arbitrary Scala function from $\texttt{T} \rightarrow \texttt{V}$, and creates an \texttt{Element[V]} that represents the application of the input element to the Scala function. Virtually any user--implemented or built--in Scala function can be incorporated into a probabilistic model using \texttt{Apply}, no matter how complex. For example, we could incorporate a function computing the centrality of a node in a social network model. 

Meanwhile, \texttt{Chain} captures sequencing of a probabilistic model by chaining the generation of one element with the generation of a subsequent element that depends on the first element. \texttt{Chain} takes a parent \texttt{Element[T]} and a function from $\texttt{T} \rightarrow \texttt{Element[V]}$. The value of a \texttt{Chain} is the value of the \texttt{Element[V]} produced by the chain function. In Bayesian network terms, a \texttt{Chain}'s first argument represents a distribution over the parent of a node, while the second argument is essentially a conditional probability distribution over the \texttt{Chain}'s output type given the parent. Using \texttt{Apply} and \texttt{Chain}, many elements can be nested together, allowing a user to quickly create complex conditional probability distributions. 

\subsection{Decision Data Structures in Figaro}

In a decision network, a decision node is a variable with an action space $V$ (i.e., its type) and a parent variable of type $T$. The parent variable represents the information available to the decision--maker at the time of decision. 
To avoid inconsistency, we use the term \emph{policy}, denoted by $\pi$, to describe a rule that tells the decision--maker what to do for any possible value of the informational parent, 
while we use \emph{decision} for the decision recommended in response to a particular value.
Given a particular value $t$ of the informational parent, the optimal decision is
\begin{equation*}
v^* = \underset{v \in V}{argmax}\, E[U | t, v]
\label{eqn:optdec}
\end{equation*}
where $U = \sum U_i$, the summation of all utility nodes in the network. Then the optimal policy for a decision node is simply
\begin{equation}
 \pi^{*} = T \rightarrow V \,|\, \pi^*(t) = v^* \, \forall\, t \in T
\label{eqn:optstrat}
\end{equation}

As Figaro is an open source, extensible language, we build decision capabilities on top of existing Figaro data structures, with no modifications to the underlying base Figaro code. Since decision nodes represent the application of a function, decision functionality should naturally utilize \texttt{Chain}: A \texttt{Decision[T,V]} is an element that inherits from \texttt{Chain} that uses a parent element of type \texttt{T} and generates a decision of type \texttt{V}. In addition, the \texttt{Decision} element also takes the range of the decision as an argument. 

\begin{figure}
\centering
\includegraphics[width=.8\columnwidth]{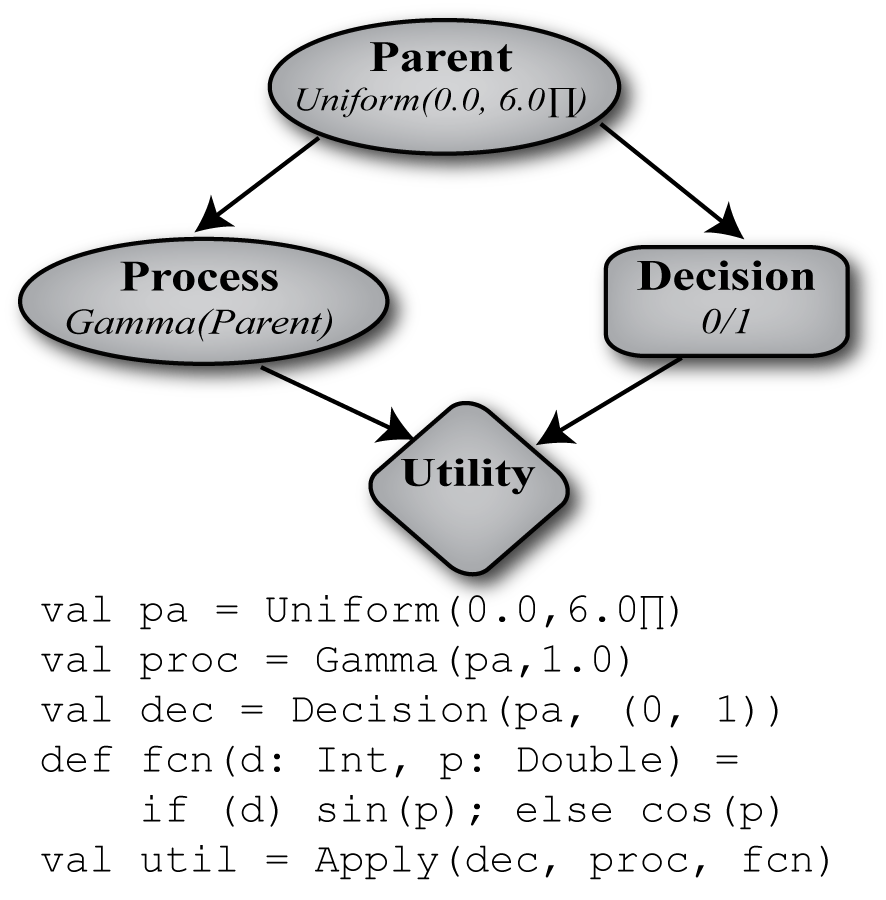}
\vspace{-4mm}
\caption[]{A decision network with continuous parents.\label{fig:BasicNetwork}}
\vspace{-2mm}
\end{figure}

For example, Fig.~\ref{fig:BasicNetwork} shows a simple decision network with continuous variables and the corresponding Figaro code. The first line generates a continuous uniform variable called \texttt{pa}, while the second line creates a gamma--distributed variable named \texttt{proc} whose shape parameter is \texttt{pa}. Line 3 defines a decision variable named \texttt{dec}, whose parent is \texttt{pa}, and whose range is the list consisting of 0 and 1. We define a function named \texttt{fcn} that takes an integer \texttt{d} and a double \texttt{p}, and returns \texttt{sin(p)} if \texttt{d} is 1, and \texttt{cos(p)} otherwise. Finally, we define a variable named \texttt{util} to be the result of applying \texttt{fcn} to \texttt{dec} and \texttt{proc}.

In decision networks, the utility of an decision is represented by a special utility node. There is no need, however, to define specialized elements to represent utilities; any variable can be used as a utility variable. All that is needed is to indicate to the reasoning algorithm which variables are utility variables. Current algorithms include decision--theoretic variants of variable elimination, importance sampling, and Metropolis--Hastings, and are described in further detail in the next section. To instantiate an algorithm for a single decision, you pass it a list of utilities and the target decision, e.g.,
\begin{Verbatim}[fontsize=\footnotesize]
 val alg = DecisionImportance(List(util),dec)
 alg.start()
 alg.setPolicy(dec)
\end{Verbatim}
The first line creates the algorithm object, the second line tells it to run and compute all the statistics necessary for decision--making, while the third line computes the optimal policy for the decision according to the collected statistics. We can then call \texttt{dec.getDecision(2.5)} to determine the best decision when \texttt{pa} is 2.5. We can also run any other standard non--decision inference algorithm to compute expected utilities under this policy. For multiple sequential decisions, another set of algorithms is provided, such as \texttt{MultiDecisionImportance}. To invoke these algorithms, you provide a list of utilities and a list of decisions.

The previous example used a simple, though infinite, data structure as a parent. Of course, the point of using probabilistic programming is to allow complex data structures and processes to be used. We now present two examples of this. 

\subsubsection{Drug Dosage Network}

\newfloat{program}{t}{}
\floatname{program}{Program}

\commentout{
\begin{program}[t]
\centering
\begin{Verbatim}[fontsize=\scriptsize]
 class Mutation(seq: Element[DNASeq], 
   time: Element[Int], mtx: Map[.]) extends
   Chain(seq, time, (t1: DNASeq, t2: Int)
    => RandWalk(t1, t2, mtx))
  
 def RandWalk(seq: DNASeq, time: Int, 
  mtx: Map[.]): Element[DNASeq] = {
   if (time == 1) return step(Constant(seq), mtx)
   else { val prev = RandWalk(seq, time-1, mtx)
    return step(prev, mtx) } } 
\end{Verbatim}
\caption{Figaro code for the random walk in \texttt{Mutation} class.\label{prog2}}
\end{program}
}

One example which we previously described concerned the dosage of some medication for a medical decision--maker. The corresponding Figaro code for the network shown in Fig.~\ref{fig:DNA} is:
\begin{Verbatim}[fontsize=\scriptsize]
 val baseProt = ``P--Y-''
 val protModel = Protein(baseProt)
 val dnaModel = DNA(protModel)
 val dnaMutate = Mutation(dnaModel, time, mtx)
 val dnaToProt = Apply(dnaMutate, mapFcn)
 val dose = Decision(protModel, (0,1,2))
 val sideEffect = Flip(0.1)
 val sideEffectUtil = Apply(sideEffect, dose, seFcn)
 val drugUtil = Apply(protModel, dnaToProt, dose, workFcn)
\end{Verbatim}
Proteins are comprised of amino acids (AA), which define the protein sequence. In line 1, the base sequence defines a set of possible protein sequences; in this case, ``P-~-Y-'', where a '-' means any of the twenty AAs. This is modeled in the \texttt{protModel} class, which defines a distribution over all possible sequences that can be generated from the base sequence (line 2). A protein is actually composed of DNA, but the mapping from proteins to DNA is a one--to--many mapping. Hence, for any given protein, we have another distribution over possible DNA sequences from the protein (line 3). On line 4, we declare an instance of a \texttt{Mutation} class, which contains elements that model the mutation of DNA over time as a random walk (using a biologically inspired transition matrix). The Figaro code for the rich protein, DNA, and mutation models is not shown. On line 5, we map the DNA back to a protein sequence using an \texttt{Apply} element. On line 6, we define the decision over three possible dosage 
levels; zero, one or two. Finally, lines 7--9 define the 
utility of the model; \texttt{sideEffectUtil} applies a negative utility depending upon the dose and side effects, and \texttt{drugUtil} produces a positive utility if the drug was effective. 

\subsubsection{Social Network}

In another example of the flexibility and expressiveness of our decision--making extensions, we implement a decision network from the social network domain. This scenario represents the decision process a company might employ to decide to provide a free product to a user in a social network, to induce her to share it with her friends. In Figaro, the program can be written as:
\begin{Verbatim}[fontsize=\scriptsize]
 val graphGen = dGraphGen(num, prob)                                    
 val rWalk = dGraphRWalk(graphGen, steps, restart)                     
 val rWalkLength = Apply(rWalk, lengthFcn)
 val freeProd = Decision(graphGen, (true,false))                  
 val prodUtil = Apply(freeProd, rWalkLength, steps, fcn)          
\end{Verbatim}

In line 1, the \texttt{dGraphGen} class models the generative process of some user's social network with a some number of nodes and probability of edge creation (these parameters could be Figaro elements as well). When a user is given a free product, they share the product with some number of other people, which can be modeled as a random walk with restart over discrete time steps. This is represented by the \texttt{dGraphRWalk} class, shown on line 2. It is more advantageous to the company if the user shares the product with many people, hence we denote the length of the random walk (distinct number of people) on line 3. The decision is invoked on line 4, where the company only observes the user's social network, but not the result of the random walk process. Finally, on line 5, the utility of the company is modeled in the function \texttt{fcn} as the number of distinct people the user shares the product with, minus the cost of the product. 
determine the best decision when only the user's social network is observed.

\subsubsection{Extensions and Limitations}

Our representation is capable of modeling decision problems with multiple decisions. For multiple decisions in sequence that obey the no--forgetting assumption, backward induction~\cite{shachter1986evaluating} can be used in conjunction with any single--decision algorithm. We do not enforce no--forgetting automatically, because we allow representation of situations that do not satisfy this assumption. As a result, our representation can also represent Limited Memory Influence Diagrams (LIMIDs)~\cite{lauritzen2001representing}, although we have not yet implemented a LIMID reasoning algorithm.

There are two main restrictions on our representation of decisions. In the declaration of \texttt{Decision} as shown above, only a single informational parent node is supported when instantiating decision objects. However, this is only a programmatic restriction and not a logical one, since the type of the parent can be any tuple of values, and Figaro supports elements over tuples. A more serious restriction is that the range of the decision must currently be discrete and finite. Support for general data types for decisions is an important direction for future work.

\section{Computing Decisions}

\subsection{Basic Framework}

The goal of our inference algorithms is to compute a policy offline that can be applied quickly online to determine the decision for any parent. 
Ideally, this decision should be as close as possible to optimal, as defined in Eq.~\ref{eqn:optstrat}. For many algorithms, the process of computing a policy for a single decision can be decomposed into two stages: computing statistics $E[U|v,t]$ for values $v$ and $t$ of the parent and decision, respectively; and using these statistics to determine a policy that takes a value for a parent and produces a decision. 

The first step is a standard probabilistic inference problem. We use Figaro's existing inference algorithms to compute expected utilities. For example, the \texttt{DecisionImportance} algorithm presented earlier uses Figaro's importance sampling to compute these expected utilities. Since these statistics may be produced by both exact and sampling inference methods, all algorithms produce a common data structure \decsamples, representing a collection of samples which maps pair $(t,v)$ of decision and parent values to a pair $(u,w)$, where $u$ is the utility observed for this sample and $w$ is the weight of the sample. This method is general: When expected utilities are computed by a non--sampling method (e.g., variable elimination), there will be at most one sample for each $(t,v)$ pair, while if the sampling method is unweighted all weights will be one.

The next step is accomplished by creating a \texttt{DecisionPolicy} (by calling \texttt{setPolicy}), which is an algorithm for determining an optimal policy. The simplest algorithm for optimal decision--making is the one used by standard decision network frameworks with discrete finite parent ranges. Given an enumerated set of parent values $t_1,...,t_n$, we can, for each parent value $t_i$, choose the $v$ (i.e., decision value) that maximizes the expected utility. Using \decsamples, the optimal policy is given by
\begin{equation*}
\pi^*(t_i) = \underset{v \in V}{argmax}\, \frac{\underset{(u,w) \in \decsamples(t_i, v)}{\sum}\!\!\!\!\!\!\!\!u*w}{\underset{(u,w) \in \decsamples(t_i, v)}{\sum}\!\!\!\!\!\!\!\!\!\!w}
\end{equation*}

This method assumes that all parent values can be enumerated; if not (due to an infinite or extremely large domain), the question arises of what to do when a parent value for which we do not have any expected utilities is encountered. Solutions are available for continuous variables, such as mixture of polynomials~\cite{shenoy2011inference}, mixture of exponentials~\cite{moral2001mixtures}, or Markov chain Monte Carlo algorithms~\cite{charnes2004multistage,chen2010simulation,garcia2007efficient}. Many of these algorithms do not work with arbitrary data structures as parents. The method proposed by~\citeauthor{chen2010simulation} could in theory be adapted to work with parents of arbitrary data structures; however, their method uses a procedure that iteratively samples utilities and updates the range of optimal decisions until convergence. Hence, for each new parent value, the complete sampling procedure must be run, creating a significant bottleneck.

Previous probabilistic programming languages have also had some decision--making capabilities, but they too do not adequately address the problem of decision--making with complex data structures. For example, IBAL~\cite{pfeffer2001ibal} used structured variable elimination to make decisions with complex probabilistic processes, but the ranges of parents were restricted to be finite and discrete. Church~\cite{goodman2008church}, meanwhile, treats decision--making as an inference problem (planning as inference) using their query function. However, Church algorithms amount to recomputing the optimal decision for every value of the parents. They do not provide a method for computing expected utility statistics and then using them to compile an offline policy. As a consequence, multiple decision networks are difficult to model in Church, as nested queries must be manually encoded by the user and significant computation is wasted determining optimal policies for multiple states of the network. 

\subsection{k--NN Algorithm}

To handle decisions with these complex parent types, we use an approximate \texttt{DecisionPolicy} that implements a k--Nearest Neighbor (k--NN) algorithm to dynamically estimate an approximately optimal policy as parent values are observed. The k--NN algorithm has been widely used in a number of domains such as search and retrieval~\cite{roussopoulos1995nearest}, classification~\cite{cover1967nearest} and estimation~\cite{cover1968estimation}. The algorithm as implemented in Figaro is relatively simple. After a decision sampling algorithm is complete, \decsamples\ is stored in a \texttt{DecisionPolicyApprox} object. When the decision is queried, upon additional simulation or inference, we estimate the optimal policy from data stored in \decsamples. First, we define a subset $\decsamples_v$of \decsamples\ for some $v \in V$ as
\begin{equation*}
\begin{split}
\decsamples_v = & \underset{t \in T}{\bigcup} \decsamples(t,v), \,\, \forall \, (t,v) \in \decsamples
\end{split}
\end{equation*}
Simply, this subset is just the set of all tuples in \decsamples\ that contain a sample from decision $v$ (with any parent value). Then the expected utility of decision $v$ with parent value $t$ is approximated as
\begin{equation*}
 \hat{E}[U|t,v] = \frac{\sum_j F(t_j)\cdot \decsamples_v(t_j,v).u \cdot \decsamples_v(t_j,v).w}{\sum_j F(t_j)\cdot \decsamples_v(t_j,v).w}
\end{equation*}
where the summations are over all entries in $\decsamples_v$. $F(t_j)$ is a function that weights the importance of each $t_j$ to $t$. Note that these nearest neighbor weights are different than $\decsamples_v(\cdot).w$, which is the weight of each utility value from the sampling algorithm. 

We want to choose a weighting function $F$ that is consistent, i.e., that ensures that $E[|\hat{E}[U|t,v] - E[U|t,v]|]\rightarrow 0$ as $n \rightarrow \infty$. For some $t \in T$ of the parent element, let $\delta(t, t')$, $t' \in T,$ be some distance measure between $t$ and $t'$. We denote $\delta_k$ as the distance between $t$ and the $k_{th}$ closest tuple of $(t',v)$ in \decsamples. Stone (\citeyear{stone1977consistent}) showed that if we set $F(t_j) = \frac{1}{k}$ when $\delta(t,t_j) \leq \delta_k$, and zero otherwise, then $F(t_j)$ produces a consistent set of weights. He also showed that a consistent weighting function can be used to estimate an approximate Bayes decision rule. Therefore, we can use the k--NN approach with confidence that it is a sound approximate reasoning algorithm for decision networks.

While our use of k--NN is conceptually very simple, two significant issues arise when using it to reason with complex data structures in a probabilistic programming framework. First, the distance function, $\delta$, must be defined for the parent type \texttt{T} of the decision. To reduce programming overhead and encourage re--use, we take advantage of several Scala features to allow users to quickly incorporate their distance functions into a decision network. 

We add an interface to Figaro called the \texttt{Distance[T]} interface. The interface consists of one function, defined as
\begin{Verbatim}[fontsize=\scriptsize]
 trait Distance[T] {def distance(that: T): Double}
\end{Verbatim}
Internally, the k--NN algorithm calls \texttt{distance} to compute nearest neighbors. Hence, any type \texttt{T} used as the parent type in an approximate policy must extend \texttt{Distance[T]} and define an implementation of the distance function. Default class extensions of simple data types (Boolean, Int, Double) that compute distance are provided, and using Scala's implicit conversion mechanism, no modifications are needed by the user. Note that any user--defined distance must return a Double value.

Furthermore, as parents of decisions may also be tuples, Figaro also contains an interface to define distance between tuples. The \texttt{Tuple} interface consists of a single function call: \texttt{reduce(List[Double]): Double}, which takes in a list of Double values and reduces it to a single value. Using the \texttt{Distance} and \texttt{Tuple} interfaces, distance for tuples is recursively computed for any combination of data types. For example, the \texttt{Tuple2} class is provided to compute the $L_2$ distance between tuples of two elements. It is declared as
\begin{Verbatim}[fontsize=\scriptsize]
 Tuple2[Distance[T1], Distance[T2]](t: (T1, T2))
  extends Tuple with Distance[(T1, T2)] with L2Norm {
   def distance(that: (T1, T2)) = 
    reduce(t._1.distance(that._1),t._2.distance(that._2)}
\end{Verbatim}
where \texttt{L2Norm} contains a declaration of \texttt{reduce} as the $L_2$ norm. Note that distance between individual types is invoked first, then \texttt{reduce} is applied. In this manner, a user can create parent tuples of disparate data types; if \texttt{Distance[T]} is defined for each type, then the distance between tuples is implicitly computed with no effort on the part of the user. These interfaces give the user a great amount of flexibility to make decisions over arbitrary classes and data types; for most models that use user--defined parent classes, extending the \texttt{Distance[T]} interface to the class and defining \texttt{distance} is all that is needed to use approximate decision policies. For instance, consider the protein sequences employed as the parent type in the dosage decision network. The class definition looks like
\begin{Verbatim}[fontsize=\scriptsize]
 class ProteinSeq(...) extends Distance[ProteinSeq] {
  def distance(that: ProteinSeq) = 
   evolDistance(this, that, distMtx) }
\end{Verbatim}
where \texttt{evolDistance} computes the evolutionary distance between proteins using the distance matrix \texttt{distMtx} (protein evolutionary distance is essentially an edit distance computation). 

The second issue relates to the design of the \texttt{DecisionPolicyApprox} data structure that contains the generated samples. Using k--NN to approximate optimal policies can result in significant performance degradation. If, after instantiating a decision network with an approximate policy, the user wishes to further sample the model, then a k--NN lookup must be performed for \textit{each} new parent sample generated. The \decsamples\ data stored in the \texttt{DecisionPolicyApprox} object may contain millions of previously generated decision samples, hence the overhead of searching for nearest neighbors in such a large data set could significantly slow down further simulation, inference and reasoning on the model.

To ameliorate this bottleneck, the data in \decsamples\ are stored in an index to facilitate quick retrieval of nearest neighbors. Two indexes are provided; a VP--tree index and a linear index. The linear index simply computes the distance between a parent value and all samples stored in \decsamples, sorts them, and returns the $k$ closest samples. The VP--tree index is a metric space partitioning index, similar to a k--d tree. Unlike a k--d tree however, the VP--tree assumes nothing about the form of the data it is indexing, and only needs a distance function between two data points to properly operate. Hence, the VP--tree index can be used for any user--defined data type with no modifications needed by the user, provided the type extends the \texttt{Distance[T]} interface. Note that the VP--tree index is a metric space tree, and therefore, only guarantees retrieval of correct nearest neighbors for metric distances. It will function for non--metric distances, though with no guarantees of correctness. Like \texttt{Distance[T]}, a standard interface for indexes is defined so that a user may implement their own index and easily integrate it into Figaro.

\section{Experiments}

\begin{figure}[t]
\centering
\subfigure{
\includegraphics[width=.80\columnwidth]{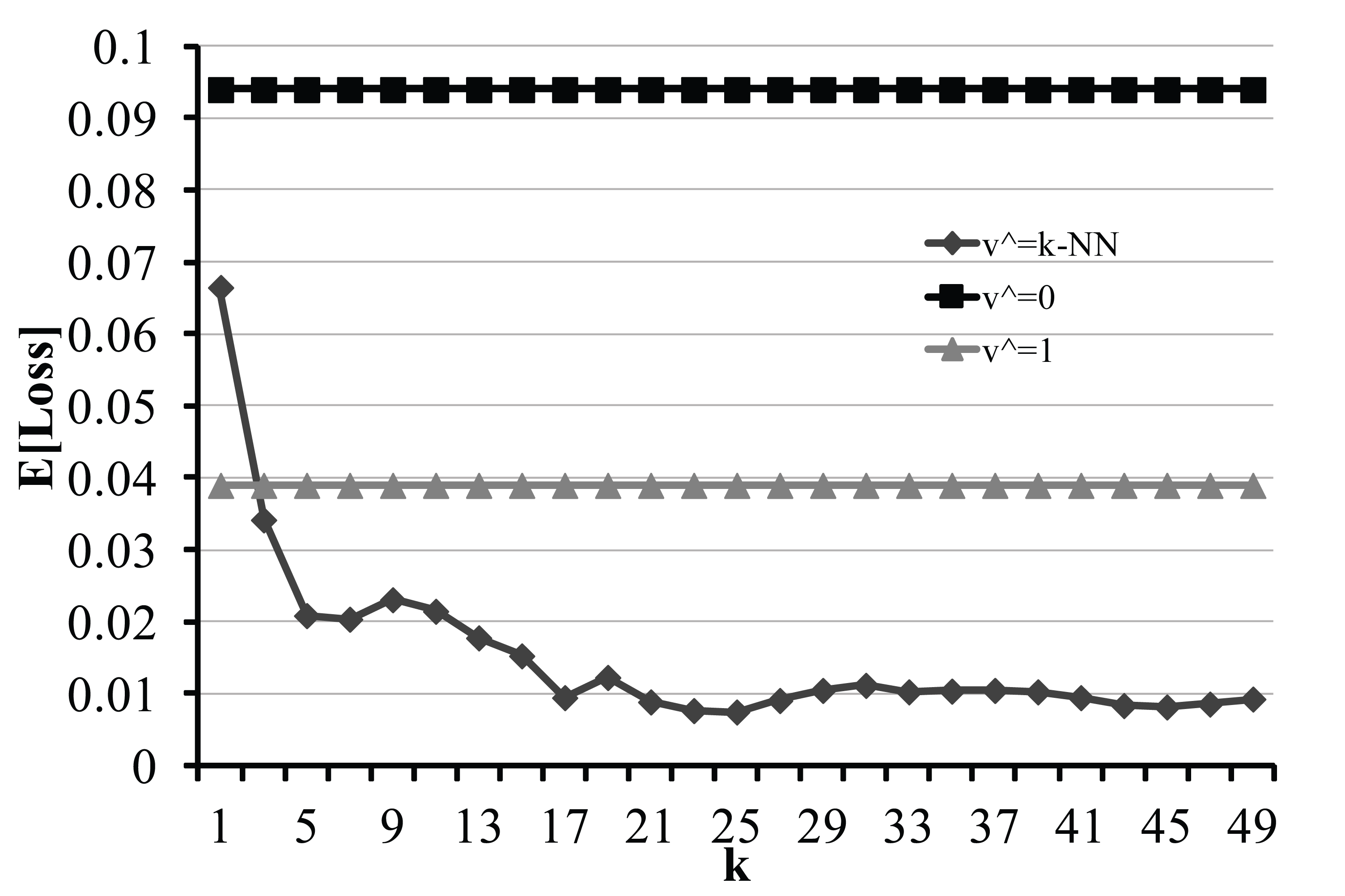}
\label{fig:BasicNetworkResultsA}
}
\subfigure{
\includegraphics[width=.80\columnwidth]{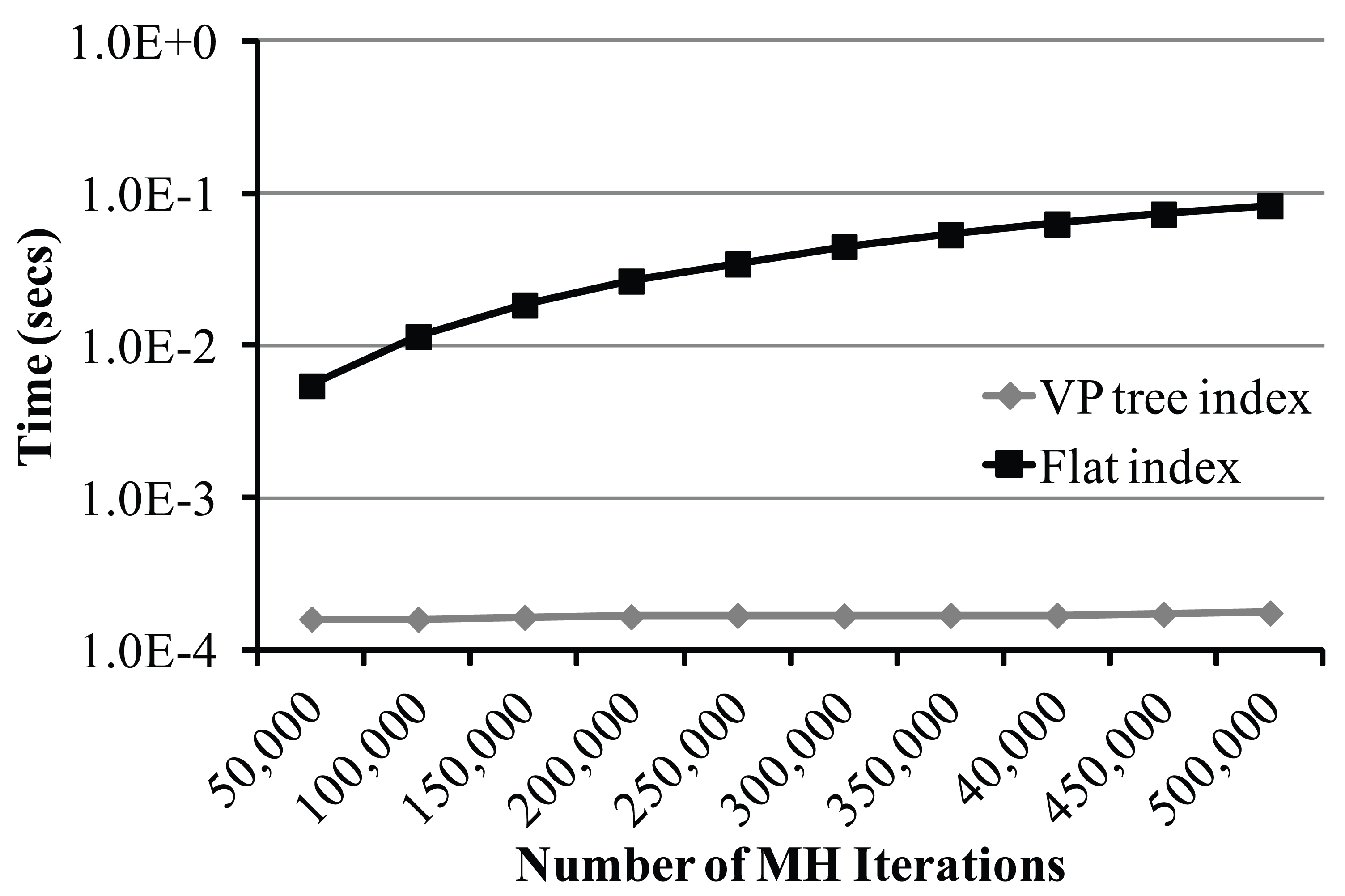}
\label{fig:BasicNetworkResultsC}
}
\vspace{-4mm}
\caption[]{The expected loss (top) and $k$--NN lookup times (bottom) on the basic model described in Fig. \ref{fig:BasicNetwork}\label{fig:BasicNetworkResults}}
\end{figure}

\begin{figure}[t]
\centering
\subfigure{
 \includegraphics[width=.80\columnwidth]{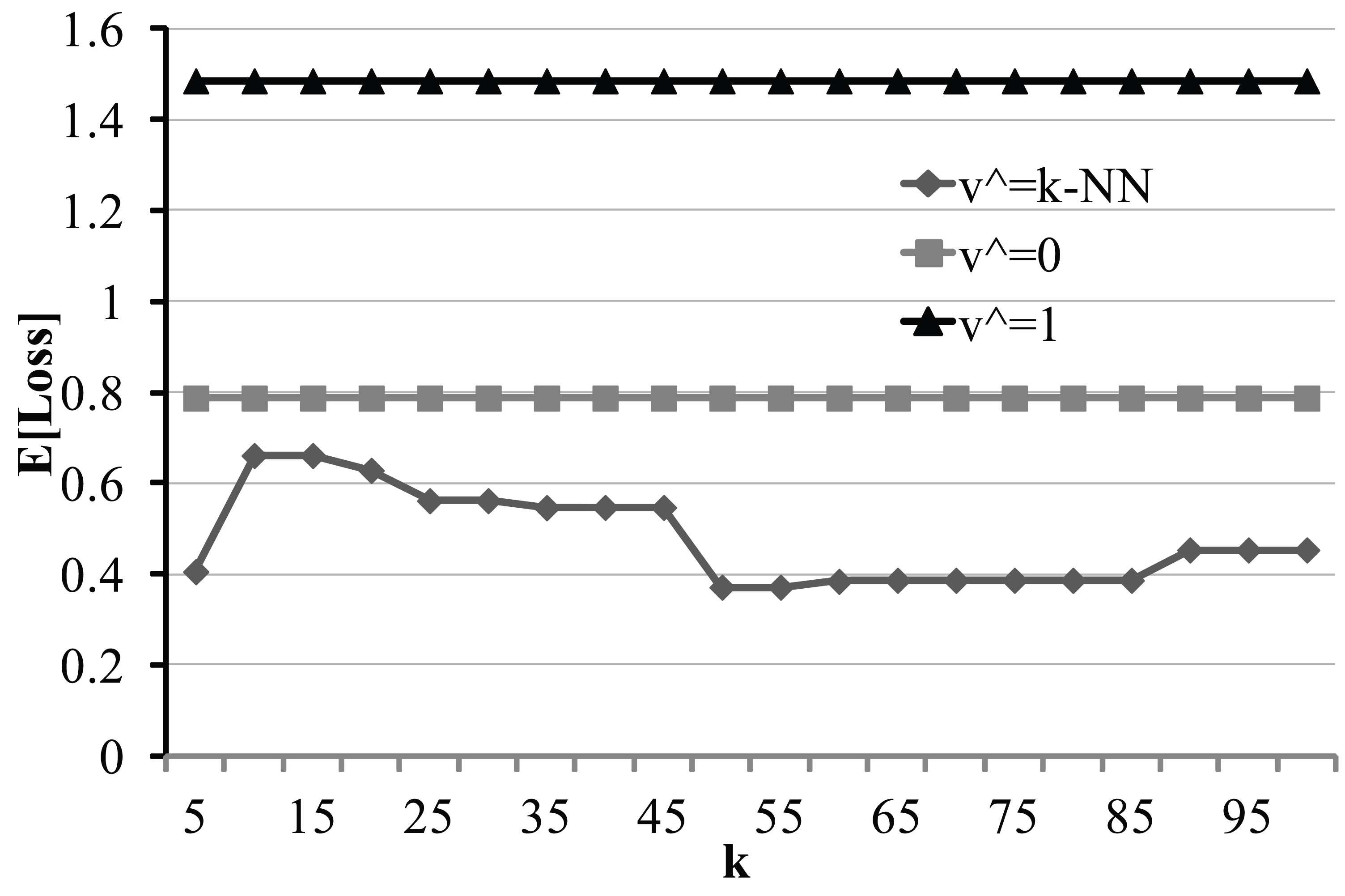}
 \label{fig:SocialNetworkResults}
}
\subfigure{
 \includegraphics[width=.80\columnwidth]{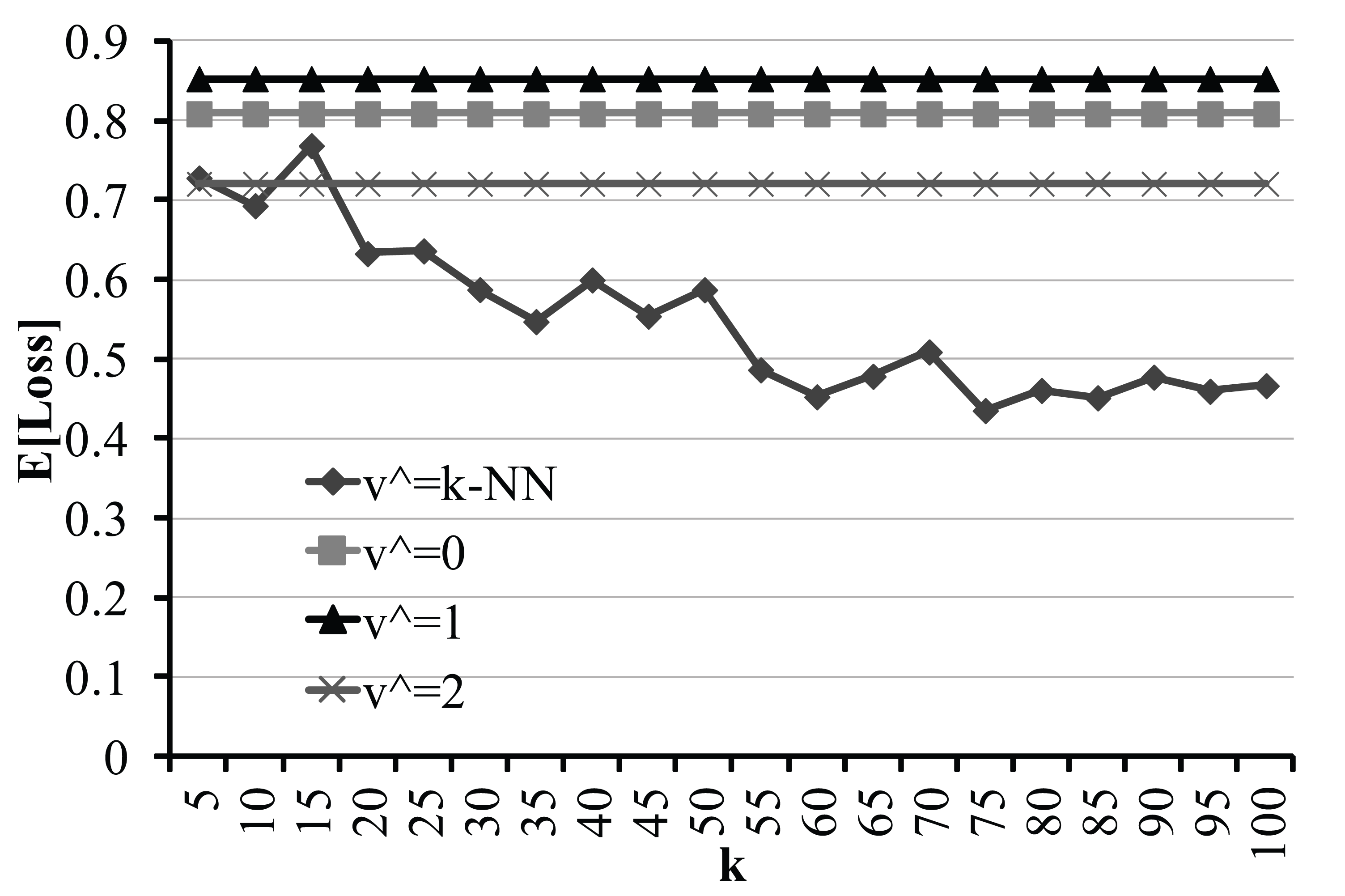}
 \label{fig:DrugResults}
}
\vspace{-4mm}
\caption[]{The expected utility loss from the social network model (top) and drug dosage model (bottom)\label{fig:OtherResults}}
\end{figure}

In this section, we perform several experiments that demonstrate the performance of the k--NN approximate decision algorithm. We first test the approximate decision algorithm on the simple decision network presented in Fig.~\ref{fig:BasicNetwork}. Although continuous parents are not the main focus of our work and other methods exist, the ability to handle continuous parents is a basic test of our algorithm's ability to handle infinite parent spaces. The testing procedure is as follows. First, the network was simulated using Metropolis--Hastings for $n$ iterations to generate the samples for \decsamples. Then, a number of testing samples were drawn from the \texttt{pa} element. For each testing sample $pa_i$, the true $E[U|pa_i,v\!=\!0]$ and $E[U|pa_i,v\!=\!1]$ were computed, as well as the k--NN approximate decision $\hat{v} \in {0,1}$. The utility loss of a decision is measured by comparing the utility obtained by the k--NN algorithm to the maximum utility obtained by an optimal decision--maker who knows 
all the final $E[U|pa_i,v]$ for each $v$. That is, it is simply  
\begin{equation*}
 Loss(\hat{v}| pa_i) = |max_v E[U|pa_i,v]  - E[U|pa_i,\hat{v}]|
\end{equation*}
The utility loss was recorded for each sample and averaged over all test samples. We also compare the k--NN loss to the loss that one would obtain if a ``static'' decision policy was employed, i.e., the decision--maker chooses the same decision all the time. Any random method of choosing a decision policy can do no better than the minimum loss achieved on these ``static'' policies.

\begin{table}[t]
  \small
  \setlength{\tabcolsep}{3pt}
  \centering
  \begin{tabular}{l | ccc|cc}
   \hline
   & \multicolumn{3}{|c|}{$\hat{v}$=k--NN} & $\hat{v}=0$ & $\hat{v}=1$ \\
   \hline
   & $k=5$ & $k=15$ & $k=25$ & & \\
   \hline
   $n=500$   & 0.023 & 0.030 & 0.038 & 0.079 & 0.046 \\
   $n=5000$  & 0.024 & 0.014 & 0.014 & 0.079 & 0.046 \\
   $n=50000$ & 0.020 & 0.012 & 0.010 & 0.079 & 0.046
  \end{tabular}
  \vspace{-1mm}
  \caption{Expected loss when $n$ is varied from $500$ to $50,000$, for three values of $k$. Note that the loss of $\hat{v}=0$ and $\hat{v}=1$ are constant as they do not depend on $k$ or $n$.\label{tbl:varyn}}  
\end{table}

The expected loss for different values of $k$, with 100 test samples, are shown in the top of Fig. \ref{fig:BasicNetworkResults}. In the figure, $n$ is fixed at $50,000$ samples. As we increase $k$, the utility loss from the approximate method steadily decreases, eventually outperforming the static decisions when $k \geq 5$.  As can be seen, even small values of $k$ can easily approximate optimal policies much better than random. In Table~\ref{tbl:varyn}, we keep $k$ constant at $5$, $15$ and $25$, and test our method on three values of $n$; $500$, $5,000$ and $50,000$ (over 10 experiments with 50 test samples each). Not surprisingly, when $n$ is small, $k=5$ has the lowest expected loss; higher values of $k$ are using too many nearest neighbors for such a small sample size. As $n$ increases, larger $k$'s can take advantage of the increased sample data to make better decisions, though eventually there are diminishing returns.

We also demonstrate that the performance of the k--NN method scales very well, as seen in the bottom of Fig. \ref{fig:BasicNetworkResults}. In this test, $k$ is kept constant at $100$, and $5,000$ queries for the approximate decision are performed using both linear and VP--tree indexes. As $n$ increases, the k--NN retrieval time using the VP--tree scales extremely well, and is orders of magnitude beyond the linear index. This demonstrates the feasibility of the approximate method even with large sets of decision samples.

To test the flexibility and expressiveness of decision--making in Figaro with complex parents, we experiment on both the social network and drug dosage examples as previously presented. Results for the social network model are shown in the top of Fig.~\ref{fig:OtherResults}, while the results for the drug dosage model are shown in the bottom of Fig.~\ref{fig:OtherResults}. The approximate decision--making policy clearly has much lower utility loss than a ``static'' decision--making policy.

\section{Conclusion}

In this work, we presented a novel probabilistic programming paradigm to represent and reason on decision networks that utilize complex information and processes. As models can be constructed that reason over arbitrary and user--defined data types, we approximate optimal decision--making policies using a k--NN algorithm. These decision--making capabilities are easy to use, and allow a user to optimize decisions over complex data types with little effort. We demonstrated that the k--NN method can accurately approximate optimal policies with little performance overhead. Our work opens the door to new applications of automated decision--theoretic reasoning.

\section{Acknowledgments}
This work was supported by DARPA contract W31P4Q-11-C-0083 and by DARPA contract FA8650-11-1-7153 under subcontract to UC Berkeley. The views expressed are those of the authors and do not reflect the official policy or position of the Department of Defense or the U.S. Government.

\begin{quote}
\begin{small}
\bibliographystyle{aaai}
\bibliography{Bibliography}
\end{small}
\end{quote}

\end{document}